\documentclass[a4paper, 10pt, conference]{ieeeconf}      
\usepackage{FG2020}

\FGfinalcopy 

\IEEEoverridecommandlockouts                              
\def\FGPaperID{136} 

\title{\LARGE \bf
Modelling the Statistics of Cyclic Activities by Trajectory Analysis on the Manifold of  Positive-Semi-Definite Matrices}

\usepackage[utf8]{inputenc} 
\usepackage[T1]{fontenc}    
\usepackage{url}            
\usepackage{booktabs}       
\usepackage{amsfonts}       
\usepackage{nicefrac}       
\usepackage{microtype}      
\usepackage{lipsum}
\usepackage{xcolor}
\usepackage{epsfig}
\usepackage{graphicx}
\usepackage{amssymb}
\usepackage{amsmath}
\usepackage[export]{adjustbox}

\author{\parbox{16cm}{\centering
    {\large Ettore Maria Celozzi$^1$, Luca Ciabini$^1$, Luca Cultrera$^1$, Pietro Pala$^1$, Stefano Berretti$^1$, \\Mohamed Daoudi$^2$, Alberto Del Bimbo$^1$}\\
    {\normalsize
    $^1$ Media Integration and Communication Center, Dept. of Information Engineering, University of Florence, Italy \\
    $^2$IMT Lille Douai, Univ. Lille, CNRS, UMR 9189 CRIStAL, F-59000 Lille, France}}
    \thanks{This work is partially funded by Fondazione Ente Cassa di Risparmio di Firenze, project \emph{PROMETEO} (ID 17807.2017.0804).}
}

\newcommand{\extended}[1]{}

\begin{document}

\ifFGfinal
\thispagestyle{empty}
\pagestyle{empty}
\else
\author{Anonymous FG2020 submission\\ Paper ID \FGPaperID \\}
\pagestyle{plain}
\fi
\maketitle

\begin{abstract}
In this paper, a model is presented to extract statistical summaries to characterize the repetition of a cyclic body action, for instance a gym exercise, for the purpose of checking the compliance of the observed action to a template one and highlighting the parts of the action that are not correctly executed (if any). 
The proposed system relies on a Riemannian metric to compute the distance between two poses in such a way that the geometry of the manifold where the pose descriptors lie is preserved; a model to detect the begin and end of each cycle; a model to temporally align the poses of different cycles so as to accurately estimate the \emph{cross-sectional} mean and variance of poses across different cycles. The proposed model is demonstrated using gym videos taken from the Internet.
\end{abstract}


\section{Introduction}
In the last decades, automatic analysis of human motion has been an active research topic, with applications that have been exploited in a number of different contexts, including video surveillance, semantic annotation of videos, entertainment, human computer interaction and home care rehabilitation, to say a few.
Differences in body proportion (size, height, corpulence), body stiffness and training, influence the way different people perform an action. Even one same person is not able to perform the same action twice, exactly replicating the same sequence of body poses in space and time. This variability makes the task of human motion analysis very challenging~\cite{barliya:2013,pollick:2001,herath:2017,Berretti:2018}.

\extended{For years, till the advent of deep CNN architectures, the approaches could be distinguished in two main classes: those that operate on motion vectors extracted from the RGB stream and those that build upon the higher level representation of body skeletons. 
These latter approaches were supported by the diffusion of low cost RGB-D cameras (such as the Microsoft Kinect) that are capable of real-time and reliable extraction of the 3D coordinates of body joints. }
Recent CNN architectures have demonstrated real-time and reliable extraction of the coordinates of body joints from RGB streams.
These advances make it possible to use a skeleton based body representation in a much broader range of operative contexts than before, not limited by the short operative range of RGB-D sensors, typically indoor and in the range of a few meters.

Many real situations do exist where computing a statistical description is of interest to characterize the execution of a particular action by a person or group of persons. For instance, extraction of such descriptors from physical activity training of athletes can be used to confirm or not the effects of some particular training exercise on the overall physical preparation of the athlete. In a different domain, the extraction of statistical descriptors from body motion can provide useful data to check the effectiveness and progress of smart rehabilitation programs. Imagine a patient that is undergoing a rehabilitation exercise at home. 
A smart assistant is capable of recognizing the patient’s actions, analyzing the correctness of the exercise, providing the patient with adequate feedback to improve the execution of the exercise, and sending statistical summaries of the rehabilitation activity to a remote server for inspection by a human therapist. 
\extended{Such a smart assistant would be greatly beneficial by 
reducing medical costs, bringing rehabilitation at home into reality.} 

Analysis of body motion can address several different tasks, such as action recognition, motion prediction, and statistical characterization of body dynamics~\cite{kong2018human}.
%
\extended{The methods proposed in this paper are mostly related to the topic of human action recognition. The literature on this subject has grown considerably in the last few years, and providing a detailed review of the many approaches to human action recognition and classification is out of the scope of this paper. The interested reader can refer to~\cite{kong2018human} for a detailed and updated survey. 
In the following, we focus the analysis on approaches that use the body skeleton as input for the recognition/classification.} 
%
One of the first approaches to analyze the trajectories of tracked body joints was presented in Matikainen \emph{et al.}~\cite{Matikainen_2009_6466} to address the task of action recognition. Their method used trajectories of tracked feature points in a bag of words paradigm. 
More recently, using the human skeleton as extracted from RGB-D images, Devanne \emph{et al.}~\cite{Devanne2015Cybernetics} proposed to formulate the action recognition task as the problem of computing a distance between trajectories generated by the joints moving during the action. An action is then interpreted as a normalized and parameterized curve in $\mathbb{R}^N$. 
In the same direction, Su \emph{et al.}~\cite{Su2014} proposed a metric that considers time-warping on a Riemannian manifold, thus allowing the registration of trajectories and the computation of statistics on the trajectories. 
Similar ideas have been developed by Ben Amor \emph{et al.}~\cite{amor2016action} on the Kendall's shape space with application to action recognition using rate-invariant analysis of skeletal shape trajectories. 
%
%
Anirudh \emph{et al.}~\cite{AnirudhTSS17} started from the framework of Transported Square-Root Velocity Fields (TSRVF). 
Based on this framework, they proposed to learn an embedding such that each action trajectory is mapped to a single point in a low-dimensional Euclidean space, and the trajectories that differ only in the temporal rate map to the same point. 
\extended{The TSRVF representation and accompanying statistical summaries of Riemannian trajectories are used to extend existing coding methods such as PCA, KSVD, and Label Consistent KSVD to Riemannian trajectories. In the experiments, it is demonstrated that such coding efficiently captures distinguishing features of the trajectories, enabling action recognition, stroke rehabilitation, visual speech recognition, clustering, and diverse sequence sampling.}
In~\cite{vemulapalli2014human, vemulapalli2016rolling}, Vemulapalli \emph{et al.} proposed a Lie group trajectory representation of the skeletal data on a product space of special Euclidean ($SE$) groups. 
\extended{For each frame, this representation is obtained by computing the Euclidean transformation matrices encoding rotations and translations between different joint pairs. The temporal evolution of these matrices is seen as a trajectory on $SE(3) \times \cdots \times SE(3)$ and mapped to the tangent space of a reference point. A one-versus-all SVM, combined with Dynamic Time Warping and Fourier Temporal Pyramid (FTP) is used for classification. One limitation of this method is that mapping trajectories to a common tangent space using the logarithm map could result in significant approximation errors.} 
\extended{In~\cite{vemulapalli2016rolling} the same authors proposed a mapping combining the usual logarithm map with a rolling map that guarantees a better flattening of trajectories on Lie groups.}
More recently, Kacem \emph{et al.}~\cite{KacemPAMI2019} proposed a geometric approach for modeling and classifying dynamic 2D and 3D landmark sequences based on Gramian matrices derived from the static landmarks. This results in an affine-invariant representation of the data. Since Gramian matrices are positive-semidefinite, the authors rely on the geometry of the manifold of fixed-rank positive-semidefinite matrices, and more specifically, to the metric investigated in~\cite{Bonnabel2009SIAM}. 
All the approaches described above rely on \textit{hand-crafted features} enabling representation of the action as a trajectory or point in some suitable manifold. Differently from these approaches, many neural network models have been proposed that rely on training for the extraction of \textit{deeply learned features}~\cite{Zhu-2016,Chao-2017,Yan-2018,Li-2019,Gao-2019}.
\extended{Recurrent Neural Networks (RNNs) and particularly Long-Short-Term Memory networks (LSTMs) have been used to perform action recognition by the analysis of sequences of skeleton poses~\cite{Zhu-2016}. However, these methods typically lose structural information when converting the skeleton data and joint connectivity into the vector-shaped input of the network. As an alternative, some approaches introduce Convolutional Neural Networks (CNNs)~\cite{Chao-2017} and Graph Convolutional Networks (GCNs)~\cite{Yan-2018,Li-2019} in the overall architecture so as to retain the structural information among joints of the skeleton.}
Although these approaches result in state-of-the-art performance~\cite{Gao-2019} on public action recognition benchmarks, it is not possible to define a formal mathematical framework to compute a valid metric on the internal, learned feature representation so as to perform a statistical analysis of the learned actions.

In this paper, a system is presented to extract statistical summaries to characterize the repetition of a cyclic body action for the purpose of checking the compliance of the observed action to a template one and highlighting the parts of the action that are not correctly performed (if any). 
Compared to previous work that focus on the task of action recognition and classification, the proposed solution aims to define a model for measuring the compliance of a user performed movement to a template movement and identify those parts of the movement that are not correctly executed. 
The proposed system relies on a Riemannian metric to compute the distance between two poses in such a way that the geometry of the manifold where the pose descriptors lie is preserved; a model to detect the begin and end of each cycle; a model to temporally align the poses of different cycles so as to accurately estimate the \emph{cross-sectional} mean and variance of poses across different cycles. 

\section{Body motion representation}\label{sec:body_motion_representation}
We assume that the observed activity corresponds to the repetition of a cyclic action and our main goal is to extract a statistics, primarily in terms of mean and deviation from this mean, to describe how the action is performed across different cycles.
The availability of these statistics makes it possible to compare the execution of a same training exercise by two different persons, aiming to identify the presence of parts of the exercise that are systematically performed in a different way by the two persons.

Adopting a skeleton based representation, the motion of a human body is represented by a sequence of body poses.
Therefore, within each cycle $c$ the action is represented by a trajectory $\Gamma_c:[0,1] \mapsto \mathbb{R}^{n}$ in the space of skeleton poses, $\Gamma_c(0)$ and $\Gamma_c(1)$ corresponding to the poses at the beginning and end of the cycle, respectively, $n$ the dimensionality of the skeleton pose representation.
To model this process, descriptors and similarity measures are defined that account for \emph{i}) how to represent a generic skeleton pose so as to guarantee invariance to a number of transforms; \emph{ii}) how to measure the dissimilarity between two poses according to the adopted skeleton pose representation; \emph{iii}) how to measure the dissimilarity between sequences of skeleton poses, so as to account for the different speed of execution of the same exercise across different cycles and across different persons. 

\subsection{Skeleton Pose Representation and Distance Measure}
The most elementary representation of a body pose corresponds to the output of the skeleton body fitting module and is in the form of a $n \times d$ matrix $\Gamma$ of coordinates of skeleton joints (where $n$ is the number of joints, and $d$ is 3 if the $x,y,z$ coordinate of each joint is considered).
This representation, that is provided by operating on RGB-D data (like with the Microsoft Kinect) or on RGB data (\emph{e.g.}, the OpenPose library~\cite{cao2018}) may prove inadequate for the task of comparing two poses, especially if we want this comparison to be invariant to some major transforms such as rotation and translation of the skeleton. 
In this case, it results much more convenient to represent a body pose through the \emph{Gram matrix} of joint coordinates:
\begin{equation}
\label{eq:gram_matrix}
G = \Gamma \; \Gamma^T \in \mathbb{R}^{n \times n} \; .
\end{equation}

\noindent
In fact, there is not a one-to-one correspondence between Gram matrices and skeleton joints matrices. Rather, any two sets of mean centered joint configurations that are the same up to an arbitrary rotation yield the same Gram matrix. Thus, the Gram matrix is representative of the equivalence class of mean centered skeleton joint matrices with respect to the rotation transform. 
By adopting a metric defined in the space of Gram matrices, it is possible to compute a rotation invariant dissimilarity measure between two skeleton poses.

Gram matrices as expressed in~\eqref{eq:gram_matrix} do not form a linear vector space, rather they span the manifold of Positive-Semi-Definite (PSD) matrices of rank $d$, the PSD cone ${\mathcal S}^+(d,n)$. As a consequence, using the Frobenius norm for computing the distance between two Gram matrices may prove inadequate, because it would not conform to the geometry of the manifold. 
The Riemannian geometry of ${\mathcal S}^+(d,n)$ was studied in~\cite{Bonnabel2009SIAM,journee2010low,vandereycken2009embedded,vandereycken2013riemannian,Massart2018,faraki2016image,meyer2011regression}.
Accordingly, a measure of the geodesic distance between two Gram matrices $G_i$ and $G_j$ is expressed as:
\begin{equation}
\label{eq:PSD-norm}
    \delta(G_i, G_j) = \left[  \mathrm{tr}(G_i) + \mathrm{tr}(G_j) - 2 \mathrm{tr}\left( \left( G_i^{\frac{1}{2}}  G_j G_i^{\frac{1}{2}} \right)^{\frac{1}{2}} \right) \right]^{\frac{1}{2}}.
\end{equation}

\subsection{Similarity between Sequences of Skeleton Poses}\label{sec:sequence_matching}
Since it cannot be expected that the action is performed exactly at the same speed in different cycles 
the poses at the same normalized time $t$ for two different cycles $c_1$ and $c_2$ of the same action may not correspond to each other. 
Thus, the computation of the \emph{cross-sectional} mean and variance of poses at the same normalized time $t$, $\left\{\Gamma_1(t), \Gamma_2(t), \ldots, \Gamma_C(t) \right\}$ would provide a fair statistical summary, with an artificially inflated variance.
Actually, to provide a more accurate measure of the {cross-sectional} mean and variance, the different trajectories should be first aligned, that is, registered in time. 
In our approach, Dynamic Time Warping~\cite{myers-1981} is used to compute the re-parametrization function $\gamma: [0,1] \mapsto [0,1]$ that best aligns the poses $\Gamma_i(t)$ and $\Gamma_j(t)$ of two generic cycles. Actaully, this poses some limitation being DTW not commutative and thus not a true metric. 

\section{The Virtual Training Coach}\label{sec:virtual-training}
The formal models defined above for computing the distance between poses and between sequences of poses have been used to design a \emph{Virtual Training Coach} application that helps practitioners, athletes and people following some physical rehabilitation therapy, to execute a gymnastic exercise with better technique. In fact, many people train themselves at home by using training coach videos on the Internet. 
As in the gym, the best way to achieve results and avoid pain is to match a good form of the exercise. 
However, at home, without a trainer or an expert that monitors the movements, it can be difficult to know if the body movement is correct.
The Virtual Training Coach aims to help users to improve their technique by comparing their activity with the same activity performed by a trainer so as to highlight those parts of the exercise (if any) that are not correctly executed. 

The application relies on the following main steps:
\begin{itemize}
\item [-] The user performs the exercise in front of a camera. The OpenPose library~\cite{cao2018} is used to extract, frame-by-frame, the coordinates of the body skeleton joints. Eventually, the sequence of all body skeleton data is stored. The same step is operated to process the video of the same exercise performed by a training coach.
\item[-] The sequence of body skeleton data represents the repetition of an elementary action several times. Thus, it is necessary to detect this elementary action so as to extract a statistics for each pose in it. This statistics is expressed in terms of mean and variance.
\item[-] The statistics extracted from the elementary actions performed by the user is compared to the statistics of the training coach to highlight those parts of the exercise performed by the user that are not correctly executed.
\end{itemize}

More details about these steps are provided in the following.

\subsection{Extraction of Body Skeleton}\label{sec:extraction-skeleton}
OpenPose~\cite{cao2018} performs pose estimation using part affinity fields, which are vectors that encode the position and orientation of limbs. 
\extended{The model is composed of a multi-stage CNN with two branches, one to learn the confidence mapping of a keypoint on an image, and the other to learn the part affinity fields. 
OpenPose is both accurate and efficient, while also scalable to process videos representing multiple people without affecting the run-time significantly. \extended{\footnote{For more information refer to the official \href{https://github.com/CMU-Perceptual-Computing-Lab/openpose}{documentation}}.} 
The joints extraction is performed for each frame of the video. 
OpenPose library gives the possibility to choose between two different skeleton models to fit the data: the \emph{BODY\_25} and the \emph{COCO} models.} 
For the experiments described in Section~\ref{sec:experiments}, the BODY\_25 model was used.
\extended{(see Figure~\ref{fig:keypoints25}). 
\begin{figure}[!ht]
\centering
\includegraphics[width=0.4\linewidth]{keypoints25}
\caption{The BODY\_25 OpenPose skeleton model.}
\label{fig:keypoints25}
\end{figure}
}
The output consists of lists containing the ordered $x$- and $y$-coordinates of model keypoints for each video frame. For each $(x,y)$ coordinate a confidence value is also provided. 
\extended{This represents a degree of reliability associated to the joint to account for occlusions and self-occlusions.}

To make the skeleton representation invariant with respect to the frame resolution, the scale of the person in the video as well as the specific biometry of the person, the coordinates of skeleton joints are normalized. In particular, joint coordinates are subject to rigid translation so as to center the hip joint to the origin of the $(X,Y)$ coordinate system. Afterwards, joint coordinates are subject to uniform scaling so as to have unitary torso length (this is the distance between the neck, the 1-st joint, and the mid-hip, the 8-th joint). 

\subsection{Detection of Action Cycles}\label{sec:cycles}
The proposed model targets the extraction of a statistical summary from the observation of an activity corresponding to the cyclic repetition, multiple times, of a base elementary action. \extended{This applies to all videos, those representing the user and those representing the training coach.} 
Hence, the sequence of skeleton poses extracted from a video is processed in order to detect this elementary action, whose repetition corresponds to the observed activity.
For this purpose, the sub-sequence of the first $T$ skeleton poses is used as template and it is compared, using a sliding window approach, to all the sub-sequences of length $T$ of the activity. 
The following \emph{cycle profile} function is computed:
\begin{equation}
\label{eq:cycle-profile}
{\cal CP}_T(t) = {\cal D} \left( S_0^{T-1}, S_t^{t+T-1} \right) \; ,
\end{equation}

\noindent
being $S_i^j$ the sequence of skeleton poses extracted from the $i$-th to the $j$-th frames and ${\displaystyle {\cal D} (\cdot, \cdot)}$ the distance function between two skeleton sequences.
\extended{As an example, the cycle profile function extracted from a sample action is shown in Figure~\ref{fig:cycleextraction}.
\begin{figure}[ht!]
\centering
\includegraphics[width=.8\linewidth]{cycle}
\caption{Sample cycle profile function.}
\label{fig:cycleextraction}
\end{figure}
}
In order to detect the period $P$ of the elementary action, local minima of the cycle profile function are detected and clustered, by minimum value, using a hierarchical clustering approach. 
The length $P$ of the elementary action, that is the period of the activity, is computed based on the mean gap between consecutive minima of the selected cluster.

\subsection{Alignment and Statistical Summaries} \label{sec:cycle_alignment}
In order to extract statistical summaries to characterize the execution of the elementary action, different cycles have to be aligned with each other.
This is accomplished by using Dynamic Time Warping to register in time the sequence of poses of the generic $i$-th cycle to those of the $(i+1)$-th cycle, as described in Section~\ref{sec:sequence_matching}. 
Eventually, each pose in the first cycle is associated with its \emph{pose-set} that groups all the corresponding poses in the next cycles.
By using DTW for the alignment, differences in terms of speed of execution of the elementary action across different cycles are coped with, enabling proper computation of the cross-sectional mean and variance of corresponding poses. 

\begin{figure*}[!ht]
\centering
\includegraphics[width=2.3cm]{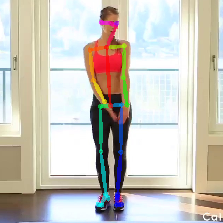}
\includegraphics[width=2.3cm, cfbox=white 2pt 2pt]{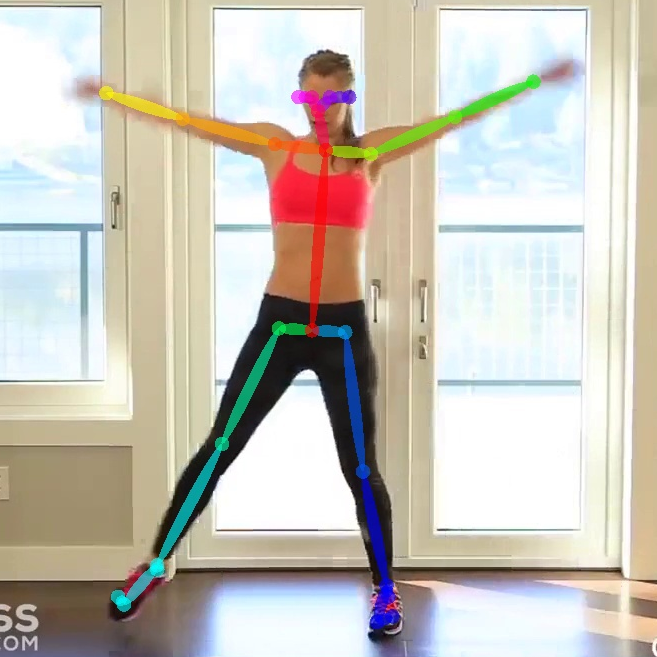}
\includegraphics[width=2.3cm]{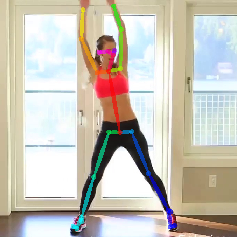}
\includegraphics[width=2.3cm, cfbox=white 2pt 2pt]{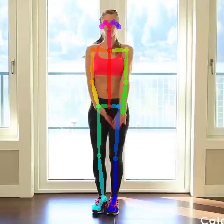}
\includegraphics[width=2.3cm]{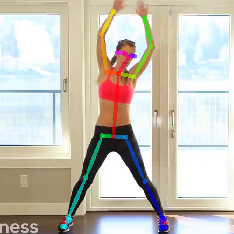}
\includegraphics[width=2.3cm]{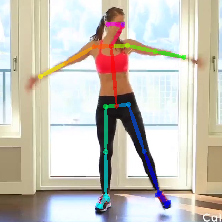}
\includegraphics[width=2.3cm]{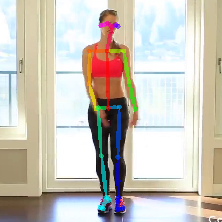}\\
\vspace{0.2cm}
\includegraphics[width=2.3cm]{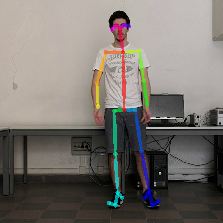}
\includegraphics[width=2.3cm, cfbox=red 2pt 2pt]{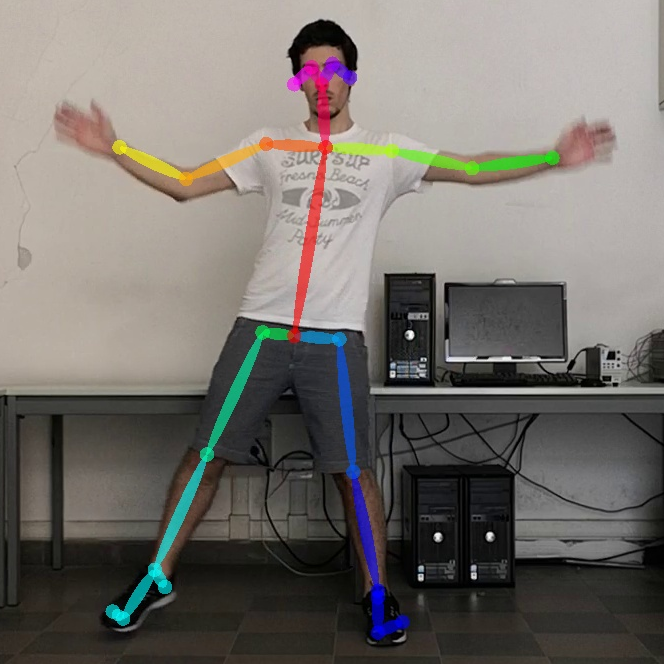}
\includegraphics[width=2.3cm]{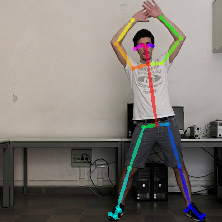}
\includegraphics[width=2.3cm, cfbox=green 2pt 2pt]{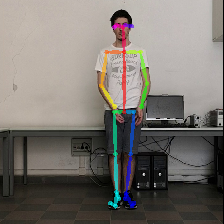}
\includegraphics[width=2.3cm]{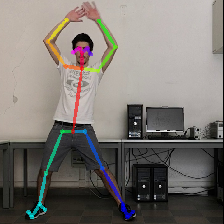}
\includegraphics[width=2.3cm]{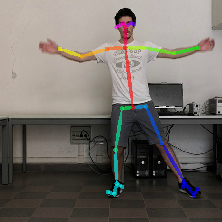}
\includegraphics[width=2.3cm]{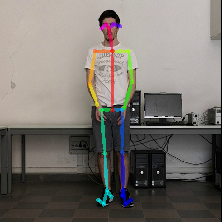}\\
\begin{flushleft}
\footnotesize \hspace{0.4cm}
pose:0 - fit:1.9\hspace{0.75cm}
pose:24 - fit:4.9\hspace{0.62cm}
pose:50 - fit:2.4\hspace{0.6cm}
pose:72 - fit:1.4\hspace{0.67cm}
pose:83 - fit:2.9 \hspace{0.35cm}
pose:92 - fit:2.2 \hspace{0.32cm}
pose:102 - fit:3.0 \normalsize
\end{flushleft}
\caption{Representative poses of one action performed by a training coach (top-row) and a user (bottom-row). The user poses characterized by the best and worst index of fit are highlighted with green and red colored frames, respectively.}
\label{fig:sample-frames-aligned}
\end{figure*}

For each pose-set the mean pose and the average deviation from this pose, using~\eqref{eq:PSD-norm}, are computed.
%
Mean and deviation extracted from the pose sets are used to compare the execution of an exercise by a user against the execution of the same exercise by the training coach.
Again, to perform this comparison the user pose sets to the training coach pose set have to be aligned.
This is accomplished by using DTW to match the sequence of mean poses extracted from the user and training coach pose sets. 
In this way, given a generic pose extracted from the user activity it is possible to identify the corresponding user pose set, its mean pose $\mu_u$ and average deviation $\sigma_u$. Value of this deviation provides a quantitative index of the ability of the user to replicate the same pose in that part of the exercise, across different cycles. Given a generic user pose set it is also possible to identify the corresponding trainer pose set, its mean pose $\mu_t$ and average deviation $\sigma_t$. Value of this deviation provides a quantitative index of the ability of the trainer to replicate the same pose in that part of the exercise, across different cycles.

To measure the accuracy of execution of the user poses across different cycles both the distances between user vs trainer mean poses as well as the deviations from these mean poses can be considered. In particular, an \emph{index of precision}, inversely proportional to the average deviation, is associated with each pose set.
Given two corresponding pose sets, for the user and the training coach, the ratio between their indexes of precision is an indicator of the relative accuracy of execution of the poses and can be used to highlight those parts of the exercise the user should focus on in order to improve the technique.
For this purpose, an \emph{index of fit} is computed for each user mean pose as $f_i = {p_i^{t}}/{p_i^{u}}$, 
%
being $p_i^{t}$ and $p_i^{u}$ the values of the index of precision for the $i$-th pose set of the training coach and the user, respectively. 
The parts of the exercise that are performed by the user with the same precision as the training coach are associated with $f_i$ values close to $1$. 
Differently, the higher the $f_i$ value the lower the user precision in the execution of the $i$-th pose.


\section{Experimental results}\label{sec:experiments}
To demonstrate the potential and effectiveness of the proposed modeling framework, we downloaded some videos of fitness exercises from YouTube and used these videos as examples of exercises performed by a training coach.
Then, for each one of these videos, we asked some users to replicate the exercise in the video.

\begin{figure}[!ht]
\centering
\includegraphics[width=0.9\columnwidth, trim={0 20 0 40}, clip]{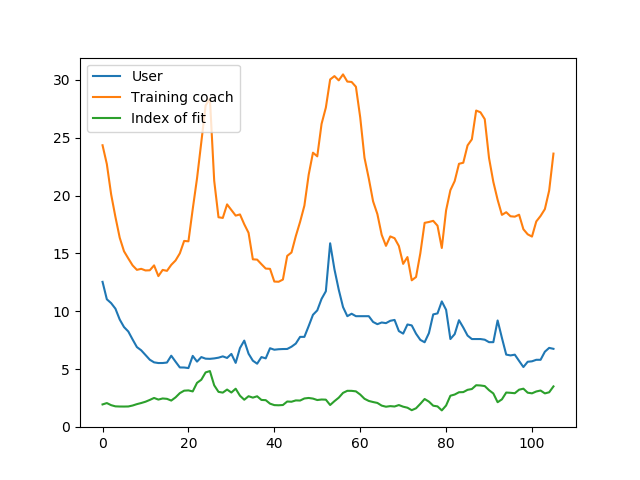}
\caption{Indexes of precision measuring the accuracy of execution for the trainer and the user shown in Figure~\ref{fig:sample-frames-aligned}. The index of fit is also shown.}
\label{fig:index-of-fig}
\end{figure}

Figure~\ref{fig:sample-frames-aligned} shows (top-row) some representative poses of one action performed by a training coach and the corresponding poses (bottom-row), after alignment, of the same action replicated by a user. 
The corresponding values of the indexes of precision and the index of fit are shown in Figure~\ref{fig:index-of-fig}.

\begin{figure}[!h]
\centering
\includegraphics[width=2.3cm]{pose24-T.png}
\includegraphics[width=2.3cm]{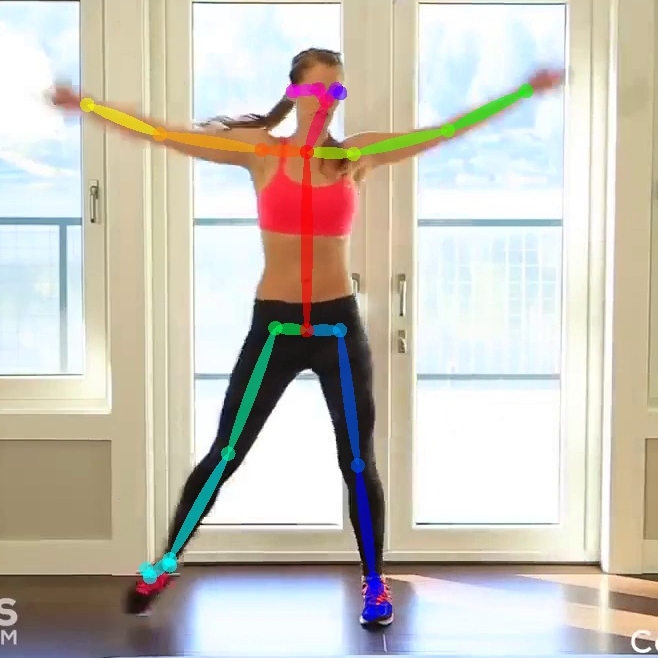}
\includegraphics[width=2.3cm]{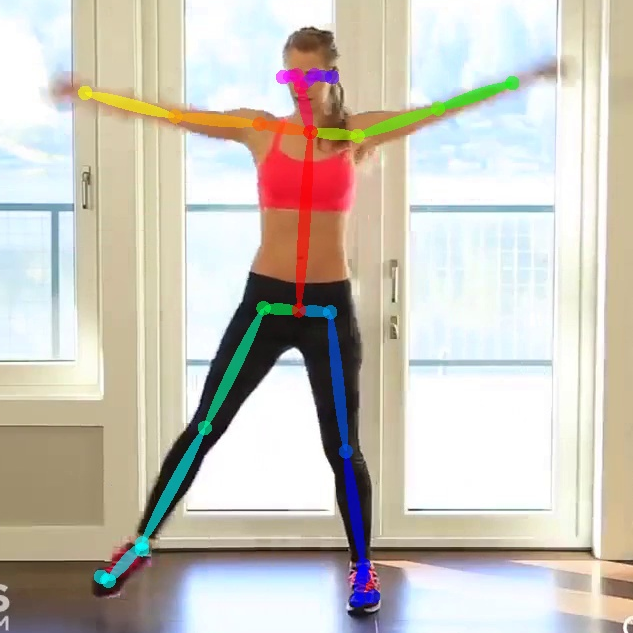}\\
\vspace{0.2cm}
\includegraphics[width=2.3cm]{pose24U.png}
\includegraphics[width=2.3cm]{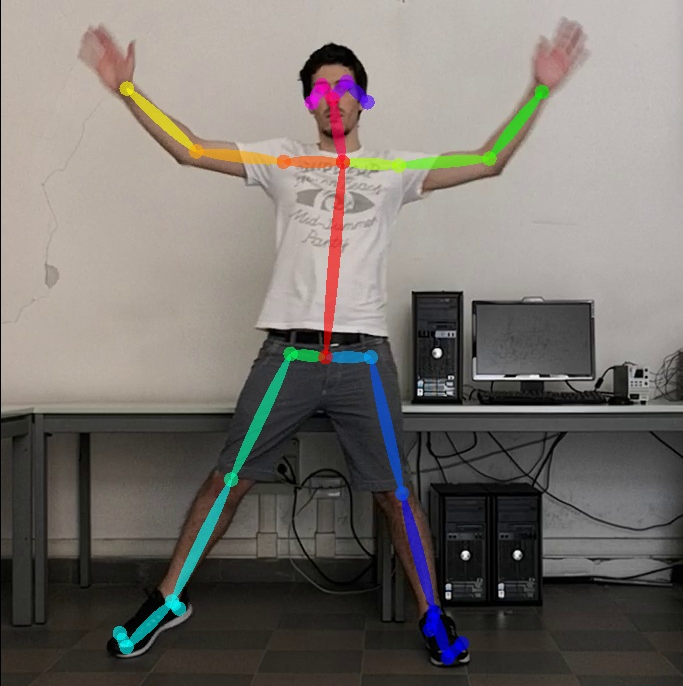}
\includegraphics[width=2.3cm]{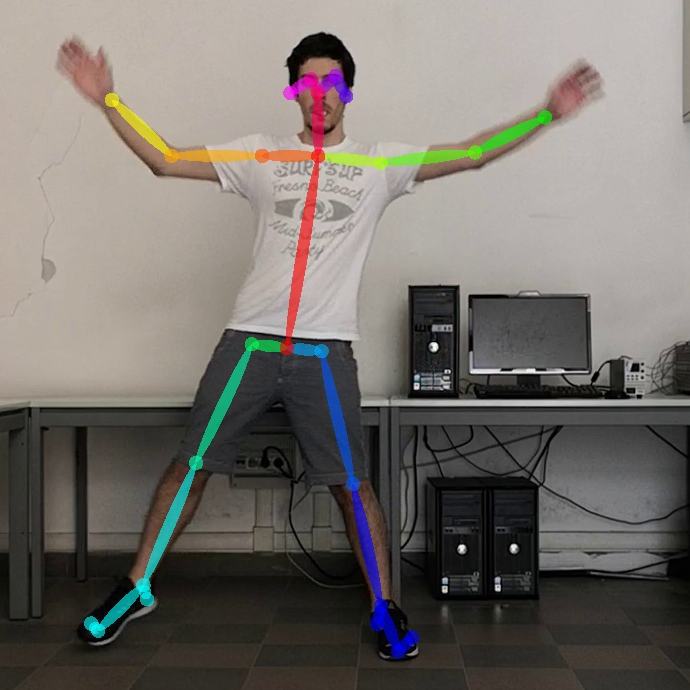}
\caption{User (bottom-row) and trainer (top-row) poses corresponding to a bad index of fit.}
\label{fig:worst-pose-set-frames}
\end{figure}

The user poses characterized by the best and worst index of fit are highlighted with a colored frame, green for the best and red for the worst one. 
By the analysis of the index of fit across the action, the user can identify the parts of the exercise to focus on in order to improve the execution.
For instance, a peak of the index of fit occurs in correspondence to the normalized time step $25$. This corresponds to the pose set with the worst index of fit: the training coach is able to replicate quite precisely the same pose across different cycles, whereas the user is not able to replicate these poses with the same precision.
Figure~\ref{fig:worst-pose-set-frames} also compares these poses for the user and the training coach: The poses of the training coach are highly similar with each other; in contrast, a higher deviation is observed among the user poses.

\vspace{-5pt}
\section{Conclusion and future work}\label{sec:conclusions}
In this paper, a model is described to extract statistical summaries for representing the accuracy of repetition of cyclic human activities. Computation of the statistical summaries is accomplished by modeling actions as trajectories in a Riemannian manifold and using a measure of the geodesic distance on this manifold to temporally align two trajectories. The proposed modeling framework is used to design a Virtual Training Coach application conceived to help users to improve their technique in the execution of a physical exercise.
Future work will target a thorough experimental evaluation of the proposed model aiming to demonstrate its effectiveness and accuracy on a representative collection of gym videos, investigating the possible gap of accuracy that is observed comparing statistics extracted from 3D and 2D skeleton data. 
\extended{Future work will target the adoption of a model to estimate the mean pose that is consistent with the geometric properties of the Riemannian manifold, such as the one described in~\cite{Bonnabel2013}.}





\end{document}